%% file: main.tex
\documentclass[10pt,twocolumn,letterpaper]{article}

\usepackage{iccv}
\usepackage{times}
\usepackage{multirow}
\usepackage{epsfig}
\usepackage{graphicx}
\usepackage{amsmath}
\usepackage{amssymb}
\usepackage{caption} 
\usepackage{cuted}
\usepackage{etoolbox}
\usepackage{booktabs}

\usepackage{amsfonts}
\usepackage{pifont}
\newcommand{\cmark}{\ding{51}}%
\newcommand{\xmark}{\ding{55}}%
\newcommand{\norm}[1]{\left\lVert#1\right\rVert}

\usepackage[table]{xcolor}
\usepackage{etoolbox}
\newtoggle{comments}
\toggletrue{comments}
\iftoggle{comments}{%
    \newcommand{\deva}[1]{{\leavevmode\color{blue}[Deva: #1]}}
    \newcommand{\achal}[1]{{\leavevmode\color{orange}[Achal: #1]}}
    \newcommand{\pavel}[1]{{\leavevmode\color{red}[Pavel: #1]}}
}{%
    \newcommand{\deva}[1]{}
    \newcommand{\achal}[1]{}
    \newcommand{\pavel}[1]{}
}

\usepackage[pagebackref=true,breaklinks=true,letterpaper=true,colorlinks,bookmarks=false]{hyperref}
\usepackage{cleveref}

\iccvfinalcopy %

\def\iccvPaperID{24} %

\newcommand{\smallsecpara}[1]{\vspace{0.5em}\noindent\textbf{#1}}

\ificcvfinal\fi
\begin{document}

\title{Learning to Track Any Object}

\author{Achal Dave\\
CMU \\
\and
Pavel Tokmakov \\
CMU \\
\and
Cordelia Schmid \\
Google Research \\
\and
Deva Ramanan \\
CMU \\
}

\maketitle

\begin{abstract}
Object tracking can be formulated as ``finding the right object in a video''. We observe that recent approaches for class-agnostic tracking tend to focus on the ``finding'' part, but largely overlook the ``object'' part of the task, essentially doing a template matching over a frame in a sliding-window. In contrast, class-specific trackers heavily rely on object priors in the form of category-specific object detectors. In this work, we repurpose category-\textit{specific} appearance models into a generic \textit{objectness} prior. Our approach converts a category-specific object detector into a category-agnostic, object-specific detector (i.e. a tracker) efficiently, on the fly. Moreover, at test time the same network can be applied to detection and tracking, resulting in a unified approach for the two tasks. We achieve state-of-the-art results on two recent large-scale tracking benchmarks (OxUvA and GOT, using external data). By simply adding a mask prediction branch, our approach is able to produce instance segmentation masks for the tracked object. Despite only using box-level information on the first frame, our method outputs high-quality masks, as evaluated on the DAVIS '17 video object segmentation benchmark.
\end{abstract}

\section{Introduction}
\label{sec:intro}

Tracking is an essential element of video analysis. Extracting spatio-temporal regions corresponding to objects from a video is not only the end goal for surveillance and video labeling~\cite{jain2016click}, but also an important intermediate representation for tasks such as action recognition~\cite{Weinzaepfel_2015_ICCV,zhang2018structured}. 

\begin{figure}[t!]
\centering
\includegraphics[width=\linewidth]{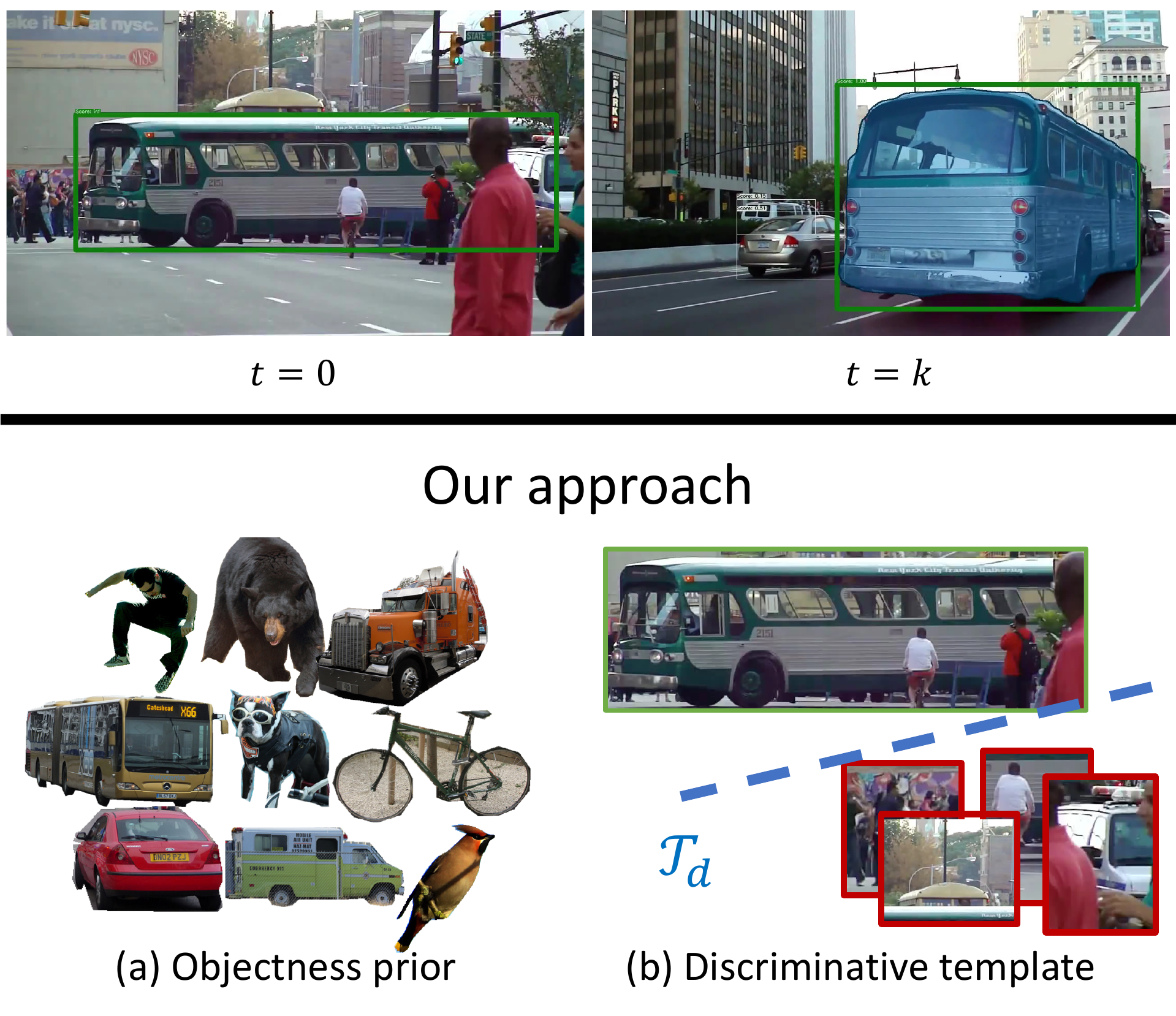}
   \caption{Objects of interest in generic, user-initialized tracking share a common set of \textit{objectness} traits. Our approach (a) learns a generic objectness prior from image-based datasets, and (b) adapts it to a specific object of interest (e.g. the bus in the top left) by computing a linear discriminator between the object and its background in closed form. This allows tracking objects through significant deformations without latching onto distractors. }
\label{fig:teaser}
\end{figure}
Unfortunately, tracking in general is notoriously challenging and potentially ambiguous. Consider the example in Figure~\ref{fig:teaser} of tracking a bus with only one side visible initially. Without prior knowledge, it is unclear whether the back side of the bus (visible in future frames) is a different viewpoint of the same bus, or a new object itself. In practice, many tracking approaches struggle to resolve such ambiguities, tending to diverge to an object part which is most similar to the initial template (e.g., the back window of the bus). Successful tracking in these scenarios necessitates \textit{object priors}. Indeed, approaches for category-specific tracking, where the tracked object categories are known before hand, heavily rely on priors in the form of category-specific object detectors~\cite{andriluka2008people,xiang2015learning,wojke2017simple,bergmann2019tracking}. By contrast, approaches for user-initialized tracking have largely eschewed such priors~\cite{bolme2010visual,bertinetto2016fully,zhu2018distractor} in the pursuit of tracking generic objects, sometimes known as model-free tracking~\cite{wu2015object,VOT_TPAMI}. However, generic objects still share a common set of \textit{objectness} traits~\cite{alexe2012measuring}. How can we operationalize this implicit constraint into a useful prior?

In this work, we repurpose category-\textit{specific} appearance models into a generic \textit{objectness} prior that can be used for category-{\em agnostic} tracking. In essence, we show that model-free tracking is far easier with better models! Doing so requires tackling two key challenges, shown in~\Cref{fig:teaser}: (1) How do we best adapt a category specific prior into a generic objectness prior? (2) How do we further adapt this generic prior to the particular instance of interest? %

To address (1), we build a joint model for category-specific object detection and category-agnostic tracking (\Cref{fig:approach}). It is based on the Mask R-CNN~\cite{he2017mask} object detection architecture. For tracking, it takes as an additional input an object template in the first frame and computes its feature embedding. 
This template is then used to compute the similarity between the object of interest and a new frame. The similarity map is in turn applied to reweight spatial features from the new frame to detect only the object of interest. Importantly, training the network jointly on image and video datasets, allows us to both capture a generic object appearance model from the diverse image data and learn to use it in a category-agnostic way for tracking.

To address (2) --  e.g., better separating the bus in~\Cref{fig:teaser} from other vehicles, such as the van on the right --  we propose a lightweight on-the-fly adaptation strategy. We compute a linear separator ($\mathcal{T}_d$ in~\Cref{fig:teaser}) between the object of interest and other objects in the first frame. This separator is computed in closed form in a fully differentiable manner, and applied in future frames to compute similarities.

An intriguing property of our proposed architecture is that it can be used both as a single-object tracker and an object detector. Moreover, by capitalizing on the mask prediction branch of~\cite{he2017mask}, we are able to train and test the same network for instance and video object segmentation. To sum up, we present a single unified approach for object detection, tracking, instance and video object segmentation.

We evaluate our model on two recent, large scale datasets for object tracking: OxUvA~\cite{valmadre2018long} and GOT~\cite{huang2018got}. The former is focused on long-term object tracking, with objects undergoing a lot of appearance variation and occlusion. In contrast, the videos in GOT are shorter, but contain diverse object categories, covering more than 560 object classes. Our method outperforms prior work on OxUvA, and outperforms state-of-the-art approaches that use external data on GOT by a large margin. Next, we show results competitive with the state-of-the-art on the LTB-35 dataset from the VOT 2018 Long Term challenge~\cite{Kristan2018a}. Finally, we validate the quality of our masks on DAVIS'17 dataset for video segmentation~\cite{Pont-Tuset_arXiv_2017}, demonstrating that our unified approach performs on par with specialized video segmentation methods that don't finetune on the test videos.

Our contributions are three-fold: (1) we incorporate an objectness prior in a generic tracker with a joint model for object detection, tracking, instance and video object segmentation; (2) we propose a lightweight strategy for computing discriminative object templates in an end-to-end fashion for efficiently handling distractors; (3) our method demonstrates state-of-the-art results on three benchmark datasets for object tracking and video object segmentation.

\input{related.tex}

\input{method.tex}

\input{exp.tex}

\section{Conclusion}
This paper introduces a novel generic object tracking approach built on top of a state-of-the-art object-detection framework. The resulting model can be trained jointly for the two tasks, effectively incorporating objectness priors into tracking. Additionally, we propose learning discriminative templates in a fully differentiable manner that encode information both about the object of interest and about the distractors, increasing the tracker's robustness. Finally, we extend our method to the related task of video object segmentation by simply adding a mask prediction branch.

Our resulting framework for tracking and video segmentation demonstrates state-of-the-art results on two recent tracking datasets (OxUvA and GOT10k), and also shows competitive performance on the DAVIS'17 benchmark for video object segmentation. We empirically show that these improvements are largely due to the generic objectness prior learned from the COCO dataset. %

{\small
\bibliographystyle{ieee}
\bibliography{main}
}

\end{document}

%% file: related.tex
\section{Related work}

\paragraph{Single object tracking.} Classical approaches for single object tracking, which requires tracking an object given a bounding box annotation in the first frame, were based on the tracking-by-detection paradigm: in many cases the detector is used to first to localize all the objects in a frame. The box corresponding to the object of interest was then selected by a discriminative classifier trained on the first frame annotation~\cite{babenko2009visual,hua2015online,kalal2012tracking}. Correlation filters were commonly used for classification due to their efficiency~\cite{bolme2010visual}. To address appearance variation, some models updated the object template over time~\cite{hua2015online,supancic2013self}. Recent approaches learn correlation filters on top of deep features~\cite{danelljan2017eco,valmadre2017end}.

Current methods for tracking largely ignore the objectness prior provided by detectors. Instead, they rely on a Siamese network architecture (initially introduced for signature verification~\cite{bromley1994signature}) adapted for tracking~\cite{bertinetto2016fully,held2016learning,tao2016siamese}. 

 \begin{figure*}[t!]
\centering
\includegraphics[width=0.8\linewidth]{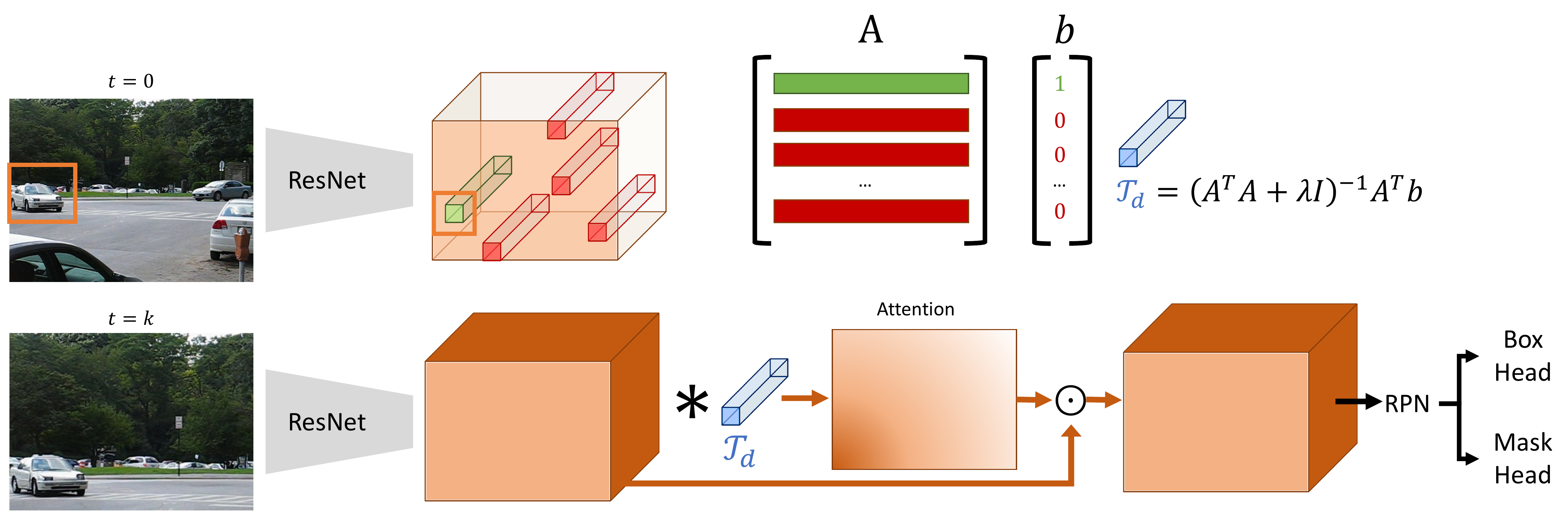}
   \caption{Overview of our approach. First, we use a state-of-the-art object detector~\cite{he2017mask} to extract features for the template image containing the object to be tracked (top left). Next, we compute a discriminative template that separates the features corresponding to the tracked object from the distractors in the first frame using linear regression (top right). Finally, attention masks computed with this template are used to reweight the feature maps of the detector to focus on the object of interest (bottom). Note that unlike standard, category-specific detectors, our box-head and mask-head output a single, category-agnostic prediction for the tracked object.}
\label{fig:approach}
\end{figure*}

Recently, there have been several attempts to introduce ideas from CNN-based detection architectures into Siamese trackers. In particular, Li et al.~\cite{li2018high} use the similarity map obtained by matching the object template to the test frame as input to an RPN-like module adapted from Faster R-CNN~\cite{ren2015faster}. Later this architecture was extended by introducing hard negative mining and template updating~\cite{zhu2018distractor}, adding a mask prediction branch~\cite{wang2019fast}, and using deeper models~\cite{li2019siamrpn++}. Our approach differs in that instead of integrating components of object detectors into a tracking pipeline in a heuristic way, we turn a state-of-the-art object detection framework into a tracker. This allows our model to fully utilize the objectness prior learned on COCO, outperforming the heuristic-based approaches significantly.

\paragraph{Video object segmentation.} Methods for video object segmentation take a precise object mask as input in the first frame and output pixel-level segmentations for the object in each frame. Early methods for this task were based on mask propagation through a graph connecting superpixels in the neighboring frames~\cite{tsai2016video,wen2015jots}. More recently, these methods have been outperformed by deep-learning based approaches, which capitalize on the success of image segmentation architectures~\cite{caelles2017one,perazzi2017learning}. In particular, they fine-tune a model trained for foreground-background segmentation using the annotation in the first frame and evaluate it on the remaining frames of the video. Some approaches also update the model using its own predictions to handle appearance variation~\cite{voigtlaender2017online}. While these methods demonstrate impressive accuracy, they remain slow due to the need to update the model during evaluation. Alternative approaches, that do not require network fine-tuning have been proposed recently~\cite{cheng2018fast,yang2018efficient,voigtlaender2019feelvos}, but remain inferior in performance.

These approaches treat video object segmentation as a problem \textit{independent} from object tracking, with the recent exception of~\cite{wang2019fast}. In contrast, we adapt the intuition from~\cite{he2017mask} that instance masks can be computed as a by-product of object detection. Our tracker with a mask prediction branch achieves competitive performance on DAVIS'17 video object segmentation benchmark without requiring mask-level supervision on the first frame.

\paragraph{Object detection.} CNN architectures for detection have brought significant progress, replacing classical methods for object detection that relied on hand-crafted features and part-based models~\cite{felzenszwalb2010object}. Early approaches ~\cite{girshick2015fast,girshick2014rich} trained CNNs to classify pre-computed object proposals. More recent approaches solve the detection problem in an end-to-end way~\cite{liu2016ssd,redmon2016you,ren2015faster}. In particular, RCNN-like architectures~\cite{he2017mask,ren2015faster} operate in a two-stage fashion: first an RPN proposes a set of boxes, and pools features from each box region. Next, separate branches classify the object and refine the box coordinates. \cite{lin2017feature} introduced feature pyramid network (FPN) to aggregate features from several network layers. Finally, Mask R-CNN~\cite{he2017mask} extended this model to instance segmentation by adding a mask prediction branch. In this work, we convert this architecture into an object tracker by introducing a lightweight discriminative template matching block before the RPN. The resulting attention map guides the RPN to propose only boxes corresponding to the object of interest. Disabling the matching component turns the model into a standard object detector.

%% file: method.tex
\section{Method}

An ideal model for tracking by detection can be described as a generic object detector that can be efficiently adapted to detect a specific object in a specific scene. In this section, we propose such an approach, shown in Figure~\ref{fig:approach}. Our model leverages advances in
standard object \textit{detection} architectures by progressively incorporating
modifications to build a state-of-the-art \textit{tracker}, while maintaining the model's detection capabilities. 

We begin by briefly describing the Mask R-CNN architecture in Section~\ref{sec:prel}. We then discuss our strategy of incorporating Siamese-like template matching into this model in a principled way in Section~\ref{sec:sim}. Next, we propose our discriminative templates that efficiently integrate information about the distractors in Section~\ref{sec:template}. Finally, we discuss our strategy for training the unified model on object detection, tracking, and video segmentation datasets in Section~\ref{sec:training}.

\subsection{Preliminaries}
\label{sec:prel}
A Mask R-CNN detector, shown in Figure~\ref{fig:approach}, consists of a backbone network (often a ResNet), a Region Proposal Network, and bounding box classification, regression and mask prediction heads. The former takes a frame as input and outputs a set of feature maps $\{C_1, C_2, C_3, C_4, C_5\}$, extracted from the respective blocks of the backbone and encoding the image with different degrees of spatial and semantic granularity. In practice, the output of the first block is discarded, due to memory constraints. The remaining feature maps are then updated via top-down lateral connections to propagate the information for the coarse but semantically rich top layers to the more spatially precise bottom layers, resulting in the final set of feature maps $\{P_2, P_3, P_4, P_5\}$ (see~\cite{lin2017feature} for details). The feature dimensionality of these maps is fixed to 256, but their spatial dimensions decrease from fine to coarse, thus the resulting architecture is referred to as Feature Pyramid Network (FPN).

An RPN is implemented as a $3 \times 3$ convolutional layer that is applied to each FPN level in a sliding window fashion, outputting an objectness score for each of the anchor boxes centered at the corresponding location. Crucially, the anchor boxes only capture various aspect ratios of the boxes, wheres scale variation is handled by the FPN. That is, a $1 \times 1 \times D$ dimensional feature at location $(x, y)$ in $P_5$ represents the largest possible object centered at  that location, whereas a feature of the same dimension at the corresponding location in $P_2$ represents the smallest possible object in centered in the same region. We use this observation to derive our scale-invariant object template in Section~\ref{sec:sim}. 

Finally, the top $k$ boxes according to the RPN score are selected, and an ROI-Pool operation is used to convert their feature representations to a fixed size. The resulting features are passed to separate bounding box classification, regression, and mask prediction branches (see~\cite{he2017mask} for details). We now describe our approach to efficiently adapting this architecture to the task of object tracking.

\subsection{Tracking as generalized object detection}
\label{sec:sim}
Given a bounding box around the object of interest in the first frame, how can we adapt the Mask R-CNN detector to only track that specific instance? 
We take inspiration from Siamese-based approaches for tracking that store an object template from the first frame and compute a similarity between the template and the test frame representation in a sliding-window fashion. Differently from those methods, instead of localizing the objects directly via template matching, we use the resulting similarity map to reweight the feature representation of an object detector. This allows us to reuse the rest of the detection architecture and train the model jointly on images and videos.

Our key observation is that every object can be represented by a $1 \times 1 \times D$ feature $\mathcal{T}$ in one FPN layer, corresponding to its scale and center location. Thus, we begin by extracting the corresponding representation for the template box in the first frame. In the standard detection training setup, Mask R-CNN assigns each groundtruth box in an image to a specific level in the feature pyramid, adding a loss that enforces that features at that scale generate a proposal around the groundtruth box. We use this same mapping to map our template box to the corresponding FPN level, and use the feature in that level that corresponds to the center of the template box. At test time, however, the scale of the object might have changed. Conveniently, we do not need to update the template to account for this, since scale variation is already handled by the FPN. Thus, we simply compute the similarity maps at all the levels of the feature pyramid via $S_i = P_i \star \mathcal{T}$, where $\star$ stands for cross-correlation.

Next, instead of directly using the resulting similarities to localize the object, we propose to instead treat them as attention maps to guide the detector. To this end, we update the original FPN representations via $P_i \leftarrow P_i \cdot S_i$, where $\cdot$ stands for the dot product. Notice that this operation simply reweights the original representation, preserving the information used by the RPN in the next stage. Thus, we can naturally capitalize on the strong objectness prior learned by detectors on COCO, as well as learn to produce objects masks for free. This re-weighted feature representation is used to generate and pool features for region proposals. The pooled, re-weighted features are finally passed through class-agnostic bounding box and mask regression heads.

At test time, our model produces multiple detections with confidence scores at every video frame. By default, we select the highest-scoring detection to construct the track, but we can make use of multiple detections by re-ranking them with external cues, such as predictions of an object dynamics model, or temporal smoothness cues (\Cref{sec:implementation}).

\subsection{Joint Detection and Tracking}
\label{sec:multitask}
The modifications described above convert a standard Mask R-CNN detector into a tracker, which can not directly be used as a standard detector. We present two modifications that allow the tracker to be trained and evaluated as a standard, image-based detection model. First, when applied to a single image, we disable the attention module. Equivalently, this can be thought of as setting the attention to a uniform value of $1$ at all pixel locations. Second, in order to output a class-specific bounding box and mask, as in standard Mask R-CNN, we instantiate a separate final layer for the box and mask regression heads for detection. Note that our model shares all parameters for detection and tracking \textit{except} this final fully connected layer. We show in \Cref{sec:abl} that training jointly for detection with tracking improves tracking accuracy, while allowing our model to operate as a powerful single-frame detector, which can be useful for identifying distractor objects during tracking.

\subsection{Discriminative Templates}
\label{sec:template}
\begin{figure}[t!]
\centering
\includegraphics[width=\linewidth]{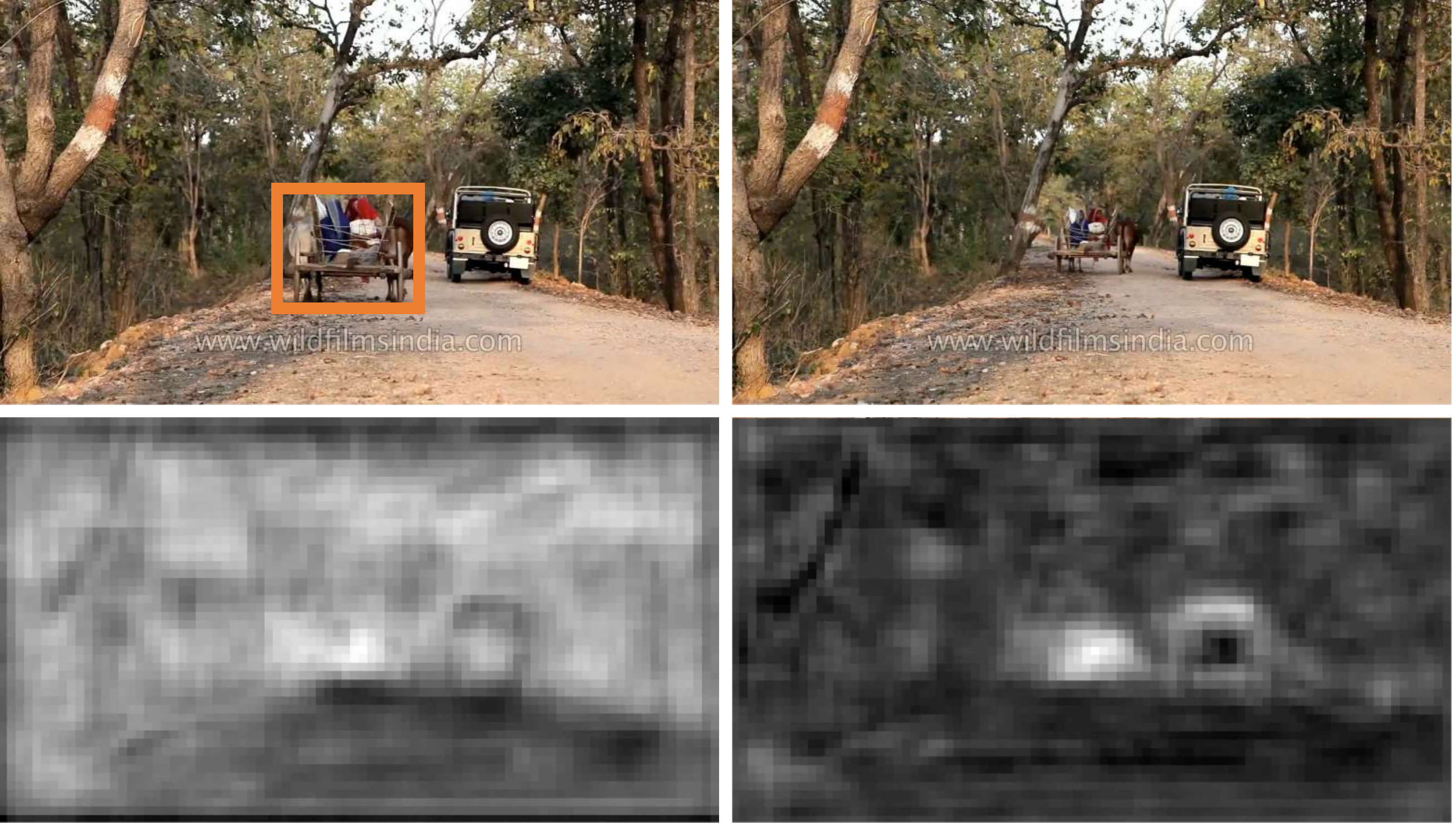}
\caption{The effect of our proposed discriminative template on an example of a video from GOT-10k dataset~\cite{huang2018got}. By simply using the center feature of the bounding box around the cart (top left), the resulting attention map (bottom left) for the test frame (top right) is not focused on the object. In contrast, our discriminative template (bottom right) results in a much better attention map.}
\label{fig:discriminative}
\end{figure}

Consider the frames from one of the videos in GOT shown in Figure~\ref{fig:discriminative} together with the corresponding similarity map $S_i$ from the appropriate FPN level of our model. The model is supposed to track the cart in this video, however, the similarity map for the test frame shown in the bottom left is not localized on the object. We now propose a simple and efficient way of learning a discriminative template, increasing the robustness of the tracker.

Recall that in the FPN a feature vector at each location encodes an object centered at that region at the corresponding scale. Thus, sampling a large enough pool of features from all the levels outside of the ground truth bounding box naturally provides us with a training set for learning a linear discriminator for the object of interest in a given video. Moreover, such a discriminator can be found efficiently in a closed form via least squares. In particular, given a template $\mathcal{T}$ and a set of negatives $N = \{\mathbf{n}_1, \mathbf{n}_2, \dots, \mathbf{n}_q\}$, we define the data matrix $A$, and the label vector $\mathbf{y}$ as follows:
\begin{equation}
A=\begin{bmatrix} 
\mathcal{T}; \mathbf{n}_1; \mathbf{n}_2; \dots; \mathbf{n}_q
\end{bmatrix},
\mathbf{y}=\begin{bmatrix} 
1; 0; 0; \dots; 0
\end{bmatrix}
\end{equation}
We then want to find a vector $\mathcal{T}_d$, which we call a discriminative template, that minimizes
\begin{equation}
\norm{A \mathcal{T}_d - \mathbf{y}}_2^2
\end{equation}
holds. A closed form solution is available via:
\begin{equation}
\mathcal{T}_d = (A^TA + \lambda I)^{-1}A^T\mathbf{y},
\end{equation}
where $I$ is the identity matrix and $\lambda$ is a regularization hyper-parameter. We then use $\mathcal{T}_d$ to compute the similarity maps in the same way: $S_i = P_i \star \mathcal{T}_d$.

Note that computing $\mathcal{T}_d$ requires only a matrix inverse and matrix multiplications, operations which are fully differentiable in the elements of $A$, and can be implemented in standard deep learning frameworks. Thus, we can backpropagate though this computation. This guides the backbone to learn a feature space where objects can be separated via a linear classifier in an end-to-end manner.

Figure~\ref{fig:discriminative} (bottom right) shows that our discriminative template indeed significantly increases the precision of the similarity maps by incorporating the information about the distractors in a principled way. We now describe how we train our unified framework on dataset for object detection, tracking, and video segmentation.

\subsection{Training}
\label{sec:training}
We first train our model on COCO for object detection~\cite{lin2014microsoft} following Mask R-CNN training~\cite{he2017mask}. We then transfer the learned objectness prior to the tracking task.

To this end we add the discriminative template computation, and attention reweighting components described above, and fine-tune it on the ImageNet VID~\cite{russakovsky2015imagenet} and YTVOS~\cite{xu2018youtube} datasets. As ImageNet Vid does not provide segmentation groundtruth, we do not use it to update the mask branch. When fine-tuning for tracking, we make three simple modifications: (1) Our training batches consist of \textit{pairs} of frames: for a batch of size $K$, we sample $K$ videos, and then sample a \textit{template} frame and a \textit{search} frame at random from the video\footnote{While we could limit the template frame to be the first frame of the video (as at test time), this would drastically reduce the diversity of our frame pairs.}; (2) We re-weight FPN features of the search frame using the feature corresponding to the template frame's bounding box; (3) Only a single, class-agnostic groundtruth box is used for training, which is the one corresponding to the tracked object in the search frame. These minor modifications allow us to maximally preserve the objectness priors learned on COCO.

%% file: exp.tex
\section{Experiments}
We begin with introducing the datasets used to train and evaluate our model, and providing the implementation details. Next we analyze the various choices made while designing our approach in Section~\ref{sec:abl}. Finally, we compare our method to the state-of-the-art in Section~\ref{sec:soa}.

\subsection{Datasets and evaluation}
\label{sec:datasets}

We use the COCO~\cite{lin2014microsoft} dataset to train our model for object detection, and ImageNet VID and YTVOS~\cite{xu2018youtube} to train the tracking module. We evaluate on two very recent, large scale tracking benchmarks: OxUvA~\cite{valmadre2018long} for long term tracking  and GOT~\cite{huang2018got} for tracking of diverse objects. In addition, we use the DAVIS'17~\cite{Pont-Tuset_arXiv_2017} dataset for video object segmentation to evaluate the quality of the masks produced by our tracker, and the LTB35 videos from the VOT 2018 long term challenge to benchmark~\cite{Kristan2018a} against prior submissions to the challenge. We describe each of these datasets in more detail in the supplementary material.

\subsection{Implementation details}\label{sec:implementation}
\smallsecpara{Network architecture and training} We use the Mask R-CNN detection framework throughout our experiments. In particular, we use the ResNet-50 FPN backbone, which achieves a useful balance between accuracy and efficiency. Our final model is trained for detection on MS COCO, as described in Sec.~\ref{sec:abl} and for bounding-box tracking on ImageNet VID and YTVOS. We will release training and evaluation code along with trained models upon acceptance.

\smallsecpara{Temporal heuristics} Prior tracking approaches rely heavily on temporal information to simplify the tracking problem. As these heuristics can obscure the improvement of the underlying matching approach, we show results using \textit{no temporal information} in Sec.~\ref{sec:abl} and ~\ref{sec:soa}, and show state-of-the-art results without heuristics on OxUvA. For completeness, we implement one simple heuristic which we ablate in Sec.~\ref{sec:abl}. At every frame $t$, our detector outputs a set of candidate detections $D_t = \{d_{1,t}, \dots, d_{k,t}\}$, along with a confidence score $c_{i,t}$ for each detection. In our standard implementation, we select the detection $d^*_t$ with the highest confidence. To incorporate temporal smoothness, we implement a simple heuristic: for each candidate box $d_{i,t}$, compute the mask intersection-over-union $j_{i,t}$ with the predicted detection $d^*_{t-1}$ at the previous frame, and update the confidence as $c_{i,t} \leftarrow \alpha c_{i,t} + (1 - \alpha) j_{i,t}$. Then, we select $d^*_t$ as the detection that maximizes this reweighted confidence. We set $\alpha=0.6$ for all of our experiments. In order to avoid latching onto distractors, we temporarily disable this smoothness component if the track is broken, i.e. the IoU between the object locations at time $t$ and $t+1$ is small ($<\alpha_\text{low}$); we re-enable the smoothness component if we maintain a smooth track for $n$ frames, i.e. a track with consecutive object locations that have IoU $>\alpha_{\text{recover}}$. We set $\alpha_\text{low}=0.1$, $\alpha_\text{recover}=0.3$, and $n=30$. We always show results both with and without this component for clarity.

\subsection{Ablation study}
\label{sec:abl}

In this section we analyze the influence of different components of our approach on the final performance. We use the \textit{dev} sets of OxUvA and GOT-10k datasets for analysis, due to their complexity and diversity. Note that OxUvA requires explicitly thresholding confidence scores in order to detect when an object is not present in the video. For ablation, we report the area under the ROC curve (i.e., TPR vs. FPR curve) to better understand the performance of ablated components across score thresholds. GOT-10k does not require setting such a threshold, so we use the standard Average Overlap metric described in \Cref{sec:datasets}.

\begin{table}
\centering
\resizebox{\linewidth}{!}{%
\begin{tabular}{lccccccc}
    \toprule
    \multirow{2}{*}{\shortstack[c]{Det\\Init?}} &
    \multirow{2}{*}{\shortstack[c]{Joint Det\\Train?}} &
    \multirow{2}{*}{$\mathcal{T}_d$} &
    \multirow{2}{*}{Smooth?} &
    \multirow{2}{*}{\shortstack[c]{OxUvA\\AUC}} &
    \multirow{2}{*}{\shortstack[c]{GOT\\AO}} \\\\\midrule
    \xmark & \xmark & center    & \xmark & 63.2 & 64.7 \\
    \cmark   & \xmark & center    & \xmark & 64.9 & 68.4 \\
    \cmark   & \cmark & center    & \xmark & 65.8 & 68.6 \\
    \midrule
    \cmark   & \cmark & mean diff & \xmark & 67.6 & 68.8 \\
    \cmark   & \cmark & mean pos  & \xmark & 69.1 & 69.1 \\
    \cmark   & \cmark & lin. reg. & \xmark & 71.1 & 69.5 \\
    \midrule
    \cmark   & \cmark & lin. reg. & \cmark & 72.1 & 73.0 \\\bottomrule
\end{tabular}
}
\caption{Evaluating the influence of different components of our approach on the OxUvA \textit{dev} and GOT-10k \textit{val} sets.See Section~\ref{sec:abl} for details.}
\label{tab:ablation}
\end{table}

We start with a baseline variant of our approach, which is trained only on videos labeled for tracking and achieves 63.2 OxUvA AUC and 64.7 GOT AO. 
Next, we evaluate the importance of object priors by pretraining our model on COCO as a generic object detector. This variant, shown in row 2, results in an 1.7\% improvement in OxUvA AUC and a 3.7\% improvement in GOT AO. 
Next, we train our model for detection and tracking jointly. As expected, this multi-task training strategy provides a modest bump on both OxUvA and GOT, leading to a model that improves in tracking while additionally being able to perform single-image detection. These improvements confirm our intuition that object priors are critical for tracking, and that the universal nature of our model is indeed helpful in transferring information from object detection datasets.

As described in Sec.~\ref{sec:template}, our framework is flexible, admitting various strategies for computing a discriminative template, $\mathcal{T}_d$. We analyze a few strategies for computing this template. In particular we compare our proposed linear regression framework (denoted as $\mathcal{T}_d=$`lin. reg.') to two simple baselines: a non-discriminative one, that simply averages several features vectors sampled from the ground truth box (denoted with $\mathcal{T}_d=$ `mean pos'), and a discriminative one that uses the difference between the means of positive and negative samples as a template (denoted with $\mathcal{T}_d=$`mean diff'). Note that these can be seen as special cases of linear regression. First, we observe that all these variants increase the model's performance, but the linear regression approach results in the largest improvement of 5.3\% OxUvA AUC and 0.9\% GOT AO. Second, the `mean diff' baseline shows the worst performance, which is counterintuitive. We attribute this result to the fact that simply subtracting the mean of the negative examples from the template leads to unstable behavior during training. In contrast, our principled approach to computing the template simplifies optimization. Finally, we show that incorporating the temporal smoothness (\Cref{sec:implementation}) provides significant improvements, particularly for short-term tracking as in GOT.

\smallsecpara{Discussion.} Performing ablations on two diverse tracking datasets allows understanding the impact of ablated components for different challenges. For example, the use of detection priors seems to be significantly more pronounced in GOT, which requires tracking a diverse set of objects, than for OxUvA. This is to be expected: while video datasets are large enough to learn priors for common objects, image-based datasets like COCO provide priors for more diverse categories. Meanwhile, our discriminative templates provide a significant improvement on OxUvA, but a more modest improvement on GOT. We attribute this to the fact that our discriminative template is able to avoid latching onto distractors when the object of interest disappears, a phenomenon that is far more common in the long-term OxUvA dataset than on GOT.

\begin{table}
\centering
\begin{tabular}{lccc}
    \toprule
    Approach                                & TPR  & TNR  & GM \\\midrule
    LCT \cite{ma2015long}                   & 22.7 & 43.2 & 31.3 \\
    MDNet \cite{nam2016learning}            & 42.1 & 0    & 32.4 \\
    TLD \cite{kalal2012tracking}            & 14.1 & 94.9 & 36.6 \\
    SiamFC + R \cite{bertinetto2016fully}   & 35.4 & 43.8 & 39.7 \\\midrule
    DaSiam~\cite{zhu2018distractor}     & 40.0 & 84.2 & 58.0 \\
    SiamMask                                & 50.4 & 88.7 & 66.9 \\
    SiamRPN++                               & 63.6 & 79.9 & 71.3 \\\midrule
    Ours w/o temporal                       & 63.2 & 79.1 & 70.8 \\
    Ours                & 65.5 & 78.2 & \textbf{71.6} \\\bottomrule
\end{tabular}
\caption{In the top half, we show the current reported state-of-the-art results on OxUvA, from~\cite{valmadre2018long}. As these methods perform poorly, we first run recent state-of-the-art trackers (DaSiam~\cite{zhu2018distractor}, SiamMask~\cite{wang2019fast}, and SiamRPN++~\cite{li2019siamrpn++}). We show that our approach significantly improves over both prior state-of-the-art, as well as these recent works.}
\label{tab:oxuva}
\end{table}

\begin{table}
\centering
\begin{tabular}{lccc}
    \toprule
    Approach            & AO  & SR \\\midrule
    DaSiam \cite{zhu2018distractor}  & 46.0 & 54.3 \\
    SiamMask~\cite{wang2019fast} & 66.8 & 78.3 \\
    SiamRPN++ ~\cite{li2019siamrpn++} & 65.8 & 77.5 \\ \midrule
    Ours w/o temporal   & 69.5 & 79.1 \\
    Ours                & \bf{73.0} & \bf{82.8} \\\bottomrule
\end{tabular}
\caption{Results on GOT-10k val set. Prior methods train only on the GOT-10k training set. For a fair comparison, we compare to DaSiam, SiamMask and SiamRPN++, which are also trained on external data. By leveraging objectness priors and detection mechanisms, our method significantly improves, likely due to the diversity of objects in GOT.}
\label{tab:got10k}
\end{table}

\subsection{Comparison to the state-of-the-art}
\label{sec:soa}
We now compare our full approach to state-of-the art methods in object tracking and video object segmentation. %

\begin{figure*}[t!]
\centering
\includegraphics[width=0.8\linewidth]{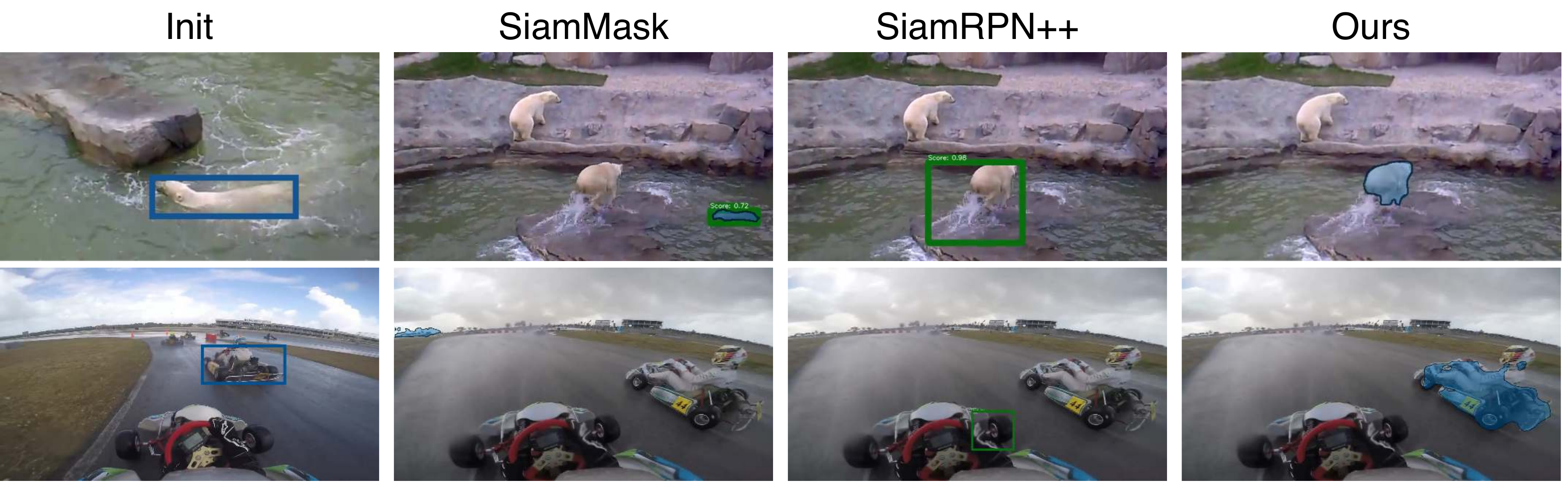}
   \caption{Qualitative comparison of our method with prior work. Top row: While SiamMask suffers from drift, and SiamRPN++ provides a loose bounding box around the object, our method provides a tight segmentation mask. Bottom row (failure case): All trackers fail to track the car of interest, which is nearly invisible in the distance. Intriguingly, while prior methods drift to background regions or parts (as with SiamMask and SiamRPN++), our method almost invariably latches onto \textit{objects}.}
\label{fig:qualitative}
\end{figure*}

\paragraph{OxUvA evaluation}
We begin by presenting comparisons on the \textit{dev} set of the OxUvA long term tracking benchmark in Table~\ref{tab:oxuva}. We compare to the state-of-the-art approaches reported in~\cite{valmadre2018long}. As the approaches reported in~\cite{valmadre2018long} perform poorly qualitatively and quantitatively, we further evaluate two more recent trackers: DaSiam~\cite{zhu2018distractor} and SiamMask~\cite{wang2019fast} on this dataset. We use their publicly available code.

As shown in Table~\ref{tab:oxuva}, our evaluation of DaSiam~\cite{zhu2018distractor}, SiamMask~\cite{wang2019fast}, and SiamRPN++~\cite{li2019siamrpn++} out-performs the methods reported in ~\cite{valmadre2018long}. Next, we evaluate our approach without using \textit{any} temporal information in the "w/o temporal" row. This variant is completely stateless, and individually performs matching on each frame of the video. By contrast, almost all prior tracking approaches use heuristic temporal smoothing to improve the performance of their models. Despite this lack of temporal information, our approach outperforms all but the very recent SiamRPN++ approach, including the recent work of~\cite{zhu2018distractor,wang2019fast}. By adding the simple temporal heuristic described in Sec.~\ref{sec:implementation}, we further improve our results by $0.8\%$, achieving state-of-the-art results. We report results on the held out \textit{test} set in supplementary.
\vspace{-4mm}

\paragraph{GOT evaluation}
To validate our conclusions above, we further evaluate our approach on GOT-10k. As prior methods evaluated on GOT-10k use only the GOT-10k training set for training, we can not fairly compare our approach to them. Instead, we use~\cite{zhu2018distractor,wang2019fast,li2019siamrpn++} as baselines, which are the best method prior to ours on OxUvA. We report results on the validation set in Table~\ref{tab:got10k}, and additionally show results on the test set in the supplementary material. As can be see from the table, we outperform all of these works by over $4\%$, which we attribute to the ability of our method to generalize to diverse object categories.

\paragraph{LTB-35 Evaluation}
We compare to state-of-the-art methods on the LTB-35 benchmark, which was used to evaluate long term tracking in the VOT 2018-LT challenge. This dataset focuses on tracking in videos over 2 minutes long on average, where the object of interest can frequently disappear and reappear in the video. We compare to state-of-the-art results in \Cref{tab:vot}, as well as SiamMask\cite{wang2019fast}. Note that prior approaches, other than \cite{wang2019fast}, provide dataset-specific hyperparameters that are tuned for this dataset. By contrast, we use a single model, a single set of hyperparameters, and a simple temporal heuristic across all datasets. Despite this, our method obtains a competitive F-measure of 61.2 on this dataset while outperforming on other datasets.

\begin{table}
    \centering
    \begin{tabular}{lccc}
        \toprule
        Approach  & P & R & F \\\midrule
        SiamMask \cite{wang2019fast}       & 64.5 & 46.8 & 54.3 \\
        DaSiam-LT \cite{zhu2018distractor} & 62.7 & 58.8 & 60.7 \\
        MBMD \cite{zhang2018learning}    & 63.4 & 58.9 & 61.0 \\
        SiamRPN++ \cite{li2019siamrpn++} & 65.0 & 61.0 & \textbf{62.9} \\\midrule
        Ours w/o temporal    & 61.0 & 56.9 & 58.9 \\
        Ours      & 61.2 & 61.2 & 61.2 \\\bottomrule
    \end{tabular}
    \caption{F-measure on LTB35 (VOT 2018-LT challenge). Unlike many prior methods, we use a \textit{single} model across all our experiments that is competitive on VOT while outperforming on other datasets.}
    \label{tab:vot}
\end{table}

\paragraph{Mask evaluation on DAVIS'17}
Finally, we evaluate our unified approach on the task of video object segmentation. To this end we use the validation set of DAVIS'17, and compare to the state-of-the-art approaches, including the ones that require finetuning the model on the test sequences. The results are presented in Table~\ref{tab:davis17-final}. We show qualitative results of our method in the supplementary material.
\begin{table}[t]
\centering
\setlength{\tabcolsep}{3pt}
\resizebox{\linewidth}{!}{%
\begin{tabular}{@{}llrrr@{\hspace{2.5em}}rr}
\toprule
\multicolumn{2}{c}{Measure}
& PReMVOS & CINM & FeelVOS & SiamMask & Ours \\
& & \cite{luiten2018premvos} & \cite{bao2018cnn} & \cite{voigtlaender2019feelvos} & \cite{wang2019fast} &  \\
\midrule
& Mask sup? & \cmark & \cmark & \cmark & \xmark & \xmark \\
& Deep FT? & \cmark & \cmark & \xmark & \xmark & \xmark \\
\addlinespace{}
\multirow{3}{*}{$\mathcal{J}$}
& Mean   &  73.9          & 67.2  & 69.1 & 54.3 & 59.2 \\
& Recall &  73.1          & 74.5  & 79.1 & 62.8 & 68.6  \\
& Decay  &  16.2          & 24.6  & 17.5 & 19.3 &  8.4 \\
\addlinespace{}
\multirow{3}{*}{$\mathcal{F}$}
& Mean   &  81.8 &  74.0         & 74.0 & 58.5 & 67.8  \\
& Recall &  88.9 &  81.6         & 83.8 & 67.5 & 76.1 \\
& Decay  &  19.5 &  26.2         & 20.1 & 20.9 & 12.0 \\
\bottomrule
\end{tabular}
}
\caption{DAVIS '17 validation results with intersection-over-union ($\mathcal{J}$) and F-measure ($\mathcal{F}$). Most prior methods require a labeled mask in the first frame (`Mask sup') or perform computationally expensive end-to-end fine-tuning per video (`Deep FT' row), our method efficiently and accurately segments objects without mask supervision.}
\label{tab:davis17-final}
\vspace{-6mm}
\end{table}

All methods in Table~\ref{tab:davis17-final}, with the exception of SiamMask and our method, require pixel-perfect segmentation in the first video frame and operate at a speed of less than 2 frames-per-second. By contrast, our method adds only a small overhead to the underlying detection model used. For our experiments, we used a ResNet-50 FPN backbone for Mask R-CNN, which led to a speed of approximately 7FPS. Despite using less supervision and computational time, our approach is competitive with dedicated video segmentation methods that use pixel-level masks in the first frame.

\paragraph{Detection evaluation on COCO} As discussed in \Cref{sec:multitask}, our model can be used as a detector at test time. Although our focus is on tracking, we find that our model outputs high quality detections, providing a COCO instance segmentation mAP of 30.5, compared to 34.4 for an equivalent standalone detector that cannot track objects.

\subsubsection{Qualitative results.} We show qualitative results in Fig.~\ref{fig:qualitative}, comparing our results with SiamMask~\cite{wang2019fast} and SiamRPN++~\cite{li2019siamrpn++}. By leveraging objectness information, our method is able to successfully localize the full extent of objects, while providing precise \textit{instance segmentation} masks for the object of interest. Further, we note an intriguing phenomenon due to our use of objectness: in failure cases, as in the bottom row of~\Cref{fig:qualitative}, our method behaves qualitatively differently from prior work. When the object of interest is not visible, our method will report a mask with low confidence. Even in these cases, our method almost invariably segments \textit{objects}, rather than drifting onto background regions or parts, as with prior methods.